\documentclass[10pt,twocolumn,letterpaper]{article}

\usepackage{cvpr}              

\definecolor{cvprblue}{rgb}{0.21,0.49,0.74}
\usepackage[pagebackref,breaklinks,colorlinks,allcolors=cvprblue]{hyperref}

\def\paperID{*} 
\def\confName{*}
\def\confYear{*}
 
\usepackage{graphicx}
\usepackage{amsmath}
\usepackage{amssymb}
\usepackage{booktabs}
\usepackage{CJKutf8}
\usepackage{xcolor}
\usepackage{makecell}
\usepackage{soul}
\usepackage{fancyhdr}

\usepackage[pagebackref,breaklinks,colorlinks]{hyperref}
\usepackage{multirow}

\usepackage[capitalize]{cleveref}
\crefname{section}{Sec.}{Secs.}
\Crefname{section}{Section}{Sections}
\Crefname{table}{Table}{Tables}
\crefname{table}{Tab.}{Tabs.}

\title{ShiftedBronzes: Benchmarking and Analysis of Domain Fine-Grained Classification in Open-World Settings}

\author{
Rixin Zhou\textsuperscript{1} \qquad
Honglin Pang\textsuperscript{1} \qquad
Qian Zhang\textsuperscript{3} \qquad
Ruihua Qi\textsuperscript{3} \qquad
Xi Yang\textsuperscript{1,2,}\textsuperscript{*} \qquad
Chuntao Li\textsuperscript{3,}\textsuperscript{*}\\
\textsuperscript{1}School of Artificial Intelligence, Jilin University\\
\textsuperscript{2}Engineering Research Center of Knowledge-Driven Human-Machine Intelligence, MoE, China \\
\textsuperscript{3}School of Archaeology, Jilin University\\
}

\definecolor{zhourixin}{RGB}{255, 20, 45}

\begin{document}
\twocolumn[{
\renewcommand\twocolumn[1][]{#1}
\maketitle
\begin{center}
    \captionsetup{type=figure}
    \centering
    \includegraphics[width=1.0\linewidth]{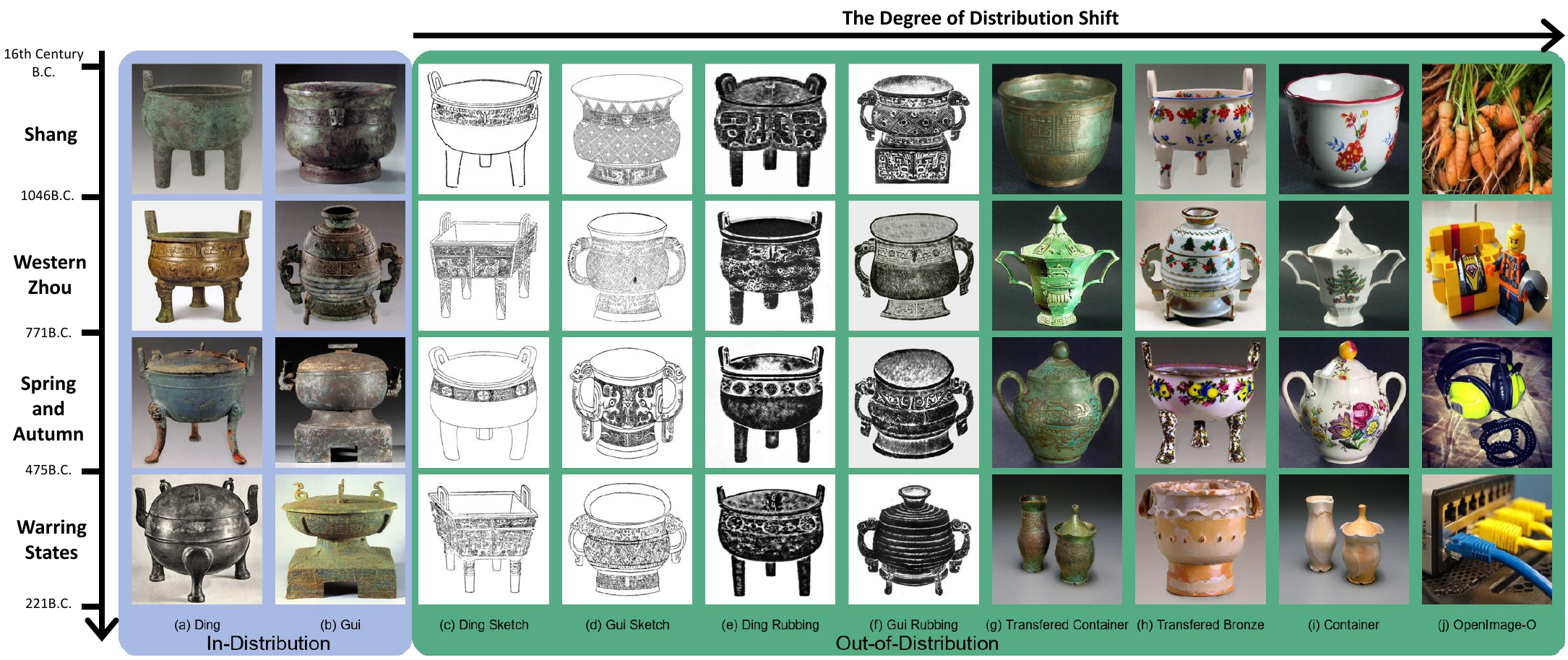}
    \captionof{figure}{Examples of our proposed dataset and a general OOD dataset (OpenImage-O~\cite{wang2022vim}). Our dataset include 2 types of ID data (a-b) and 7 types OOD data (c-i). (a) and (b) are typical images of bronze Ding and Gui from four dynasties (Shang, Western Zhou, Spring and Autumn, Warring States), which together form the ID data for the bronze ware dating task. (c-f) are sketch images and rubbing images of Ding and Gui. (g) and (h) are generated images by applying a zero-shot material transfer model to bronze ware and container images. (i) are container images sourced from the ImageNet-21K~\cite{ridnik2021imagenetk} dataset. (j) are examples from OpenImage-O. In terms of bronze ware dating scenarios, the distribution shift between the ID data and the OOD data increases from left to right within the green region.}
  \label{fig1: teaser}
\end{center}
}]

\begin{abstract}
\vspace{-20pt}

In real-world applications across specialized domains, addressing complex out-of-distribution (OOD) challenges is a common and significant concern. In this study, we concentrate on the task of fine-grained bronze ware dating, a critical aspect in the study of ancient Chinese history, and developed a benchmark dataset named ShiftedBronzes. By extensively expanding the bronze Ding dataset~\cite{zhou2023multi}, ShiftedBronzes incorporates two types of bronze ware data and seven types of OOD data, which exhibit distribution shifts commonly encountered in bronze ware dating scenarios. We conduct benchmarking experiments on ShiftedBronzes and five commonly used general OOD datasets, employing a variety of widely adopted post-hoc, pre-trained Vision Large Model (VLM)-based and generation-based OOD detection methods. Through analysis of the experimental results, we validate previous conclusions regarding post-hoc, VLM-based, and generation-based methods, while also highlighting their distinct behaviors on specialized datasets. These findings underscore the unique challenges of applying general OOD detection methods to domain-specific tasks such as bronze ware dating. We hope that the ShiftedBronzes benchmark provides valuable insights into both the field of bronze ware dating and the and the development of OOD detection methods. The dataset and associated code will be available later.

\end{abstract}
\thispagestyle{fancy} 
      \fancyhead{} 
      \fancyfoot{} 
      \fancyfoot[L]{\footnotesize *Corresponding authors}
      \renewcommand{\headrulewidth}{0pt} 
      \renewcommand{\footrulewidth}{1pt} 
      \fancyfootoffset[R]{-14cm}

\vspace{-20pt}
\section{Introduction}

In practical applications within the open world, specialized domains often encounter complex out-of-distribution (OOD) issues. Most existing fine-grained visual classification (FGVC) methods, including those for bronze ware dating, are trained based on the closed-world assumption~\cite{krizhevsky2012imagenet, he2015delving}, where the test data is assumed to be drawn independent and identically distributed (i.i.d.) from the same distribution as the training data. In contrast with general OOD benchmarks~\cite{yang2022openood, anonymous2024oodrobustbench, zhao2022ood} on ImageNet classification, the domain fine-grained benchmarks are limited. Studies on OOD benchmarks have been proposed for specific areas of expertise, such as materials property prediction~\cite{omee2024structure}, drug discovery~\cite{ji2022drugood} and medical image~\cite{cao2020benchmark}. However, these domain-specific OOD benchmarks focus more on the development of their own fields, lacking consideration of how domain-specific data could advance OOD detection methods. Additionally, there is a lack of benchmark research on the archaeological dating of bronze wares in open-world settings.

To fill this gap and advance the development of bronze ware dating in open-world scenarios, we constructed a comprehensive benchmark ShiftedBronzes for bronze ware dating by significantly expanded the existing bronze Ding dataset~\cite{zhou2023multi}. The ShiftedBronzes dataset consists of seven data components, identified as Ding, Gui, Ding sketch, Gui sketch,  Ding rubbing, Gui rubbing, transferred container, transferred bronze, container. Figure~\ref{fig1: teaser} illustrates examples from our proposed dataset alongside samples from a general OOD dataset, OpenImage-O~\cite{wang2022vim}. In the bronze ware dating task, Ding and Gui data act as in-distribution (ID) data, while the remaining data types are OOD. There are a total of 6,208 color, sketch, and rubbing images of Ding and Gui. All expanded bronze ware data have been thoroughly annotated with expert knowledge. There are 51,023 images for the transferred container, transferred bronze, and container, respectively.

The characteristics of our dataset are as follows. As illustrated in the green OOD region of Figure~\ref{fig1: teaser}, the distribution shift between the bronze ware ID data and the OOD data gradually increases from left to right. These data correspond to different types of distribution shift encountered in the bronze ware dating scenario. Among the OOD data, sketch and rubbing images are specialized types of data commonly encountered in archaeological contexts. Sketches capture the shape and decoration of bronze wares, while rubbings use ink to transfer three-dimensional details onto a flat surface. These two types of specialized images exhibit smaller distribution shifts compared to other OOD data types, which makes them more similar to the bronze ware ID data in terms of appearance and structure. Generated bronze images simulate the material properties of real bronze ware and are used to represent unknown types of bronze wares in the context of bronze ware dating. Similarly, generated container images depict craft items that resemble the shapes of bronze wares, posing challenges in distinguishing them from genuine bronze ware. Given that both Ding and Gui function as containers, modern container images further complicate the task of bronze ware dating by introducing additional visual noise.

Based on ShiftedBronzes and five general OOD datasets, we benchmark six representative FGVC methods in the bronze ware dating task and eighteen representative OOD detection methods in the bronze ware OOD detection task. Our evaluation of the FGVC methods in the bronze ware dating task aligns with settings specifically designed for bronze Ding dating~\cite{zhou2023multi}. The OOD detection methods we selected can be classified into three categories: post-hoc-based, pretrained vision language models (VLMs)-based, and generation-based methods.

The OOD detection experiments conducted on ShiftedBronzes not only reaffirm established conclusions from general OOD data but also provide new insights specific to OOD detection in specialized datasets. Compared to general OOD data, current methods face greater challenges when detecting domain-specific OOD samples with small distribution shifts. Among the methods compared, VLM-based methods consistently outperform both post-hoc and generation-based methods. By analyzing the performance differences among various VLM-based methods, we uncover the factors contributing to their superior performance in both domain-specific and general OOD detection, while also revealing distinct ID-OOD correlations in specialized versus general domains. The top-performing post-hoc methods highlight critical factors influencing OOD detection, suggesting directions for future improvements in post-hoc techniques. Additionally, pre-trained diffusion model-based methods demonstrate a clear advantage in detecting OOD samples from the sketch and rubbing categories compared to other OOD data types.






The main contributions are summarized below:
\begin{itemize}
    \item We propose ShiftedBronzes, a comprehensive benchmark designed for evaluating bronze ware dating and OOD detection methods in scenarios characterized by diverse real-world domain distribution shifts.
    
    \item We benchmark six representative FGVC methods for the bronze ware dating task and classify eighteen representative OOD detection methods into three categories for evaluation in the bronze ware OOD detection task.
    
    \item A comprehensive analysis of the experimental results led to conclusions that contribute to both bronze ware dating research and the development of OOD detection methods.

\end{itemize}


\label{Related Work}
\section{Related Work}
\subsection{OOD Benchmark}
Currently, numerous OOD benchmarks have been proposed to assess and compare the performance of various algorithms, models, or systems. OpenOOD~\cite{yang2022openood} categorizes OOD datasets into different difficulties, and establishes multiple benchmarks to provide a comprehensive evaluation of various OOD detection methods. Benchmark studies on OOD detection have been initiated for specific areas of expertise. Omee et al.~\cite{omee2024structure}conducted a benchmarking analysis to evaluate the robustness of structure-based graph neural networks in the context of predicting properties of OOD materials. Ji et al.~\cite{ji2022drugood} present a systematic OOD dataset curator and benchmark for AI-aided drug discovery. Cao et al.~\cite{cao2020benchmark} undertake a benchmarking analysis of prevalent OOD detection methods across three pivotal areas of medical imaging. Studies on the robustness of OOD detection~\cite{anonymous2024oodrobustbench, zhao2022ood, mao2023coco} benchmark the performance of OOD methods under various distribution shifts by introducing perturbation variables into images. However, existing domain-specific OOD benchmarks primarily focus on advancing their respective fields, without addressing how domain-specific data can enhance OOD detection methods. Furthermore, there is a notable absence of benchmark research on the archaeological dating of bronze wares within open-world scenarios.

\subsection{OOD Detection Methods}
OOD detection is a crucial task in computer vision, vital for ensuring robust model performance in real-world applications. This challenge has attracted considerable research, with a particular emphasis on post-hoc methods and those based on Vision-Language Models (VLMs). These approaches are favored due to their practical advantages, including ease of implementation and relevance to real-world use cases. Among the studies on post-hoc OOD detection methods, Hendrycks et al.~\cite{hendrycks2017a} proposed a baseline OOD detection method using the maximum softmax probability (MSP) as an ID score. Subsequent studies have investigated alternative, efficient indicators to differentiate between ID and OOD samples, such as gradient-based methods~\cite{huang2021importance, liang2018enhancing}, mahalanobis distance-based method~\cite{lee2018simple}, energy-based functions~\cite{liu2020energy}, gram matrix-based method~\cite{pmlr-v119-sastry20a} and weight sparsification-based method~\cite{sun2022dice}. Among the studies on VLM-based OOD detection methods, Ming et al.~\cite{ming2022delving} have extended the MSP score to VLM, examining the effects of softmax probabilities and temperature scaling on OOD detection performance. LoCoOp~\cite{miyai2024locoop} performs OOD regularization that utilizes the portions of CLIP~\cite{radford2021learning} local features as OOD features during training. CLIPN~\cite{wang2023clipn} fine-tuned CLIP to enable it to output negative prompts to assess the probability of a OOD concept. ID-like~\cite{bai2024id} explore ID-like OOD samples in the vicinity space of ID samples leveraging CLIP and utilizing prompt learning framework to identify ID-like outliers to further leverage the capabilities of CLIP for OOD detection.

Generation-based OOD detection methods also represent a significant research focus. GANs have been used to synthesize unknown samples from known classes, aiding in distinguishing between known and unknown samples for open set recognition~\cite{zhou2021learning, kong2021opengan} and OOD detection~\cite{sricharan2018building, vernekar2019out}. Some methods prioritize reconstruction quality from input data~\cite{schlegl2017unsupervised} or estimate data density via generative models~\cite{Serr2020Input}. Recent research has utilized diffusion models for detecting OOD and novelty instances~\cite{mirzaei2022fake, graham2023denoising}. For example, DIFFGUARD~\cite{gao2023diffguard} applies diffusion models to detect OOD instances with semantic discrepancies by conditioning generation on both input images and semantic labels. Additionally, Du et al.~\cite{du2024dream} leverages a diffusion model to produce high-resolution, realistic outliers, enhancing a model’s capacity to identify unknown inputs during deployment.

\subsection{Bronze Ware Dating}
Bronze ware dating task aims at determining the specific historical age to which the bronze ware belongs. The chemical and physical properties of metals are used to locate the exact year of a bronze ware~\cite{DatingArchaeologicalBronzeArtifacts,ElectrochemicalCharacterizationAndDatingOfArchaeological, ProvenanceOfZhouDynastyBronzeVessels, CharacterizationOfCorrodedBronzeDing}. Zhou et al. ~\cite{zhou2023multi} consider the dating of bronze Ding as a fine-grained classification problem and examine several FGVC methods~\cite{P2PNet, YourFL, HRN, NTS, SPS} to address the bronze ware dating task. They also collected a dataset of bronze Ding and developed a multi-granularity bronze ware dating model. Furthermore, the research in ~\cite{li2023ai} has established an online mini-program platform designed to recognize the eras of bronze Ding. However, current methods are based on the i.i.d. assumption, making it challenging to handle OOD input samples in open-world scenarios.


\section{ShiftedBronzes Benchmark}
\label{ShiftedBronzes benchmark}
Our goal is to establish a comprehensive benchmark specifically tailored to real-world scenarios in bronze ware dating. To achieve this, we have curated two types of bronze ware data and seven types of OOD data that reflect common distribution shifts encountered in this domain.

The bronze ware data includes color images of Ding and Gui, which serve as the ID data used to train FGVC methods for bronze ware dating. This data is split into training, validation, and test sets, as detailed in Section~\ref{Data Preparation}. Additionally, we include sketch and rubbing images, which are professional representations of bronze ware forms, patterns, and inscriptions, captured using specialized techniques. These images are commonly used in the context of bronze ware dating. The "Transferred container data" consists of images simulating other categories of bronze wares or counterfeit items, while "Transferred bronze data" mimics artificial craft items that resemble bronze wares in shape. Containers, sharing functional similarities with bronze wares, are often mistaken for them, which makes distinguishing between these categories challenge in the context of OOD detection.

\begin{table}[h]
\centering
\caption{Statistics of bronze ware, sketch, rubbing and transferred container data.}

\label{tab: data Statistics}
\resizebox{\linewidth}{!}{  
\begin{tabular}{ccccccccccccc}
\hline\hline
\multicolumn{1}{c|}{\multirow{2}{*}{Era}}     & \multicolumn{2}{c|}{Shang}         & \multicolumn{3}{c|}{Western Zhou}        & \multicolumn{3}{c|}{Spring and Autumn}   & \multicolumn{3}{c|}{Warring States}     & \multirow{2}{*}{Total} \\
\multicolumn{1}{c|}{}                         & Early & \multicolumn{1}{c|}{Late}  & Early & Mid  & \multicolumn{1}{c|}{Late} & Early & Mid  & \multicolumn{1}{c|}{Late} & Early & Mid & \multicolumn{1}{c|}{Late} &                        \\ \hline
\multicolumn{13}{c}{Bronze Ware Data (ID)}                                                                                                                                                                                               \\ \hline\hline
\multicolumn{1}{c|}{Ding}                     & 93    & \multicolumn{1}{c|}{945}   & 801   & 410  & \multicolumn{1}{c|}{264}  & 298   & 189  & \multicolumn{1}{c|}{215}  & 85    & 78  & \multicolumn{1}{c|}{141}  & 3519                   \\ \hline
\multicolumn{1}{c|}{Gui}                      & 8     & \multicolumn{1}{c|}{355}   & 732   & 550  & \multicolumn{1}{c|}{352}  & 138   & 16   & \multicolumn{1}{c|}{27}   & 8     & 5   & \multicolumn{1}{c|}{1}    & 2192                   \\ \hline
\multicolumn{13}{c}{Sketch, Rubbing and Transferred Container Data (OOD)}                                                                                                                                               \\ \hline
\multicolumn{1}{c|}{Ding Sketch and Rubbing} & 0     & \multicolumn{1}{c|}{67}    & 51    & 18   & \multicolumn{1}{c|}{15}   & 4     & 4    & \multicolumn{1}{c|}{2}    & 2     & 0   & \multicolumn{1}{c|}{8}    & 171                    \\ \hline
\multicolumn{1}{c|}{Gui Sketch and Rubbing}  & 0     & \multicolumn{1}{c|}{46}    & 113   & 72   & \multicolumn{1}{c|}{78}   & 14    & 2    & \multicolumn{1}{c|}{0}    & 0     & 0   & \multicolumn{1}{c|}{1}    & 326                    \\ \hline
\multicolumn{1}{c|}{Transferred Container}      & 918   & \multicolumn{1}{c|}{11718} & 13805 & 8166 & \multicolumn{1}{c|}{5544} & 3924  & 1854 & \multicolumn{1}{c|}{2196} & 846   & 756 & \multicolumn{1}{c|}{1296} & 51023                  \\ \hline
\end{tabular}
}
\end{table}

\subsection{Bronze Ware, Sketch and Rubbing Data}
\paragraph{Data Collection.}
We divide the bronze Ding dataset~\cite{zhou2023multi} into three categories: color, sketch, and rubbing images of bronze ware. To expand these datasets, we collect and annotate 2518 Gui images from both five published archaeology books and four websites, which are also organized into color, sketch, and rubbing categories. The quantitative statistics of Ding and Gui dataset are presented in Table~\ref{tab: data Statistics}.

\begin{figure}[t]
\centering
  \includegraphics[width=1\linewidth]{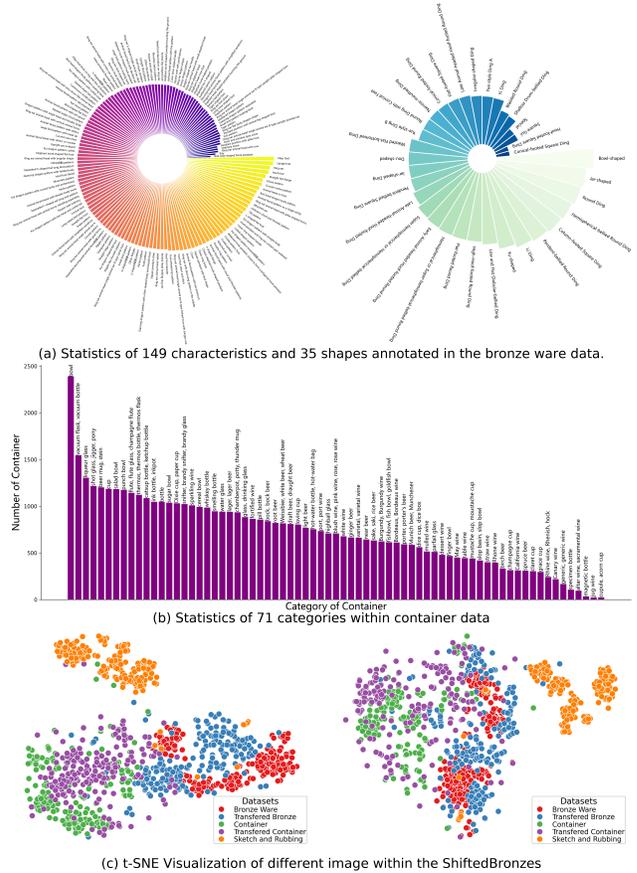}
  \caption{(a) Statistical of expert-annotated knowledge within the bronze ware dataset. (b) Statistics of categories within container data. (c) Feature visualization of different image within the ShiftedBronzes. Features are extracted using ResNet-50 (left) and ViT-B-16 (right) models, both pre-trained on the ImageNet-1K dataset, followed by dimensionality reduction via t-SNE.}
  \label{fig: mergedDatasetInformation}
\end{figure}

\vspace{-6pt}
\paragraph{Data Annotation.}
All bronze Gui have been annotated with comprehensive expert knowledge, including era (4 course-grained dynasties and 11 fine-grained periods), attributes (35 shapes and 149 characteristics with bounding boxes), literature, locations of excavation, and current exhibition museums. The numbers of shapes and characteristics labels are shown in Figure~\ref{fig: mergedDatasetInformation} (a). Because many dating results are controversial, we re-argue the era of each artifact through discussions with three bronze experts. The collection and labelling of data were carried out by an archaeologist and eight archaeology assists, who took approximately eight months to complete.

\begin{figure}[t]
\centering
  \includegraphics[width=1\linewidth]{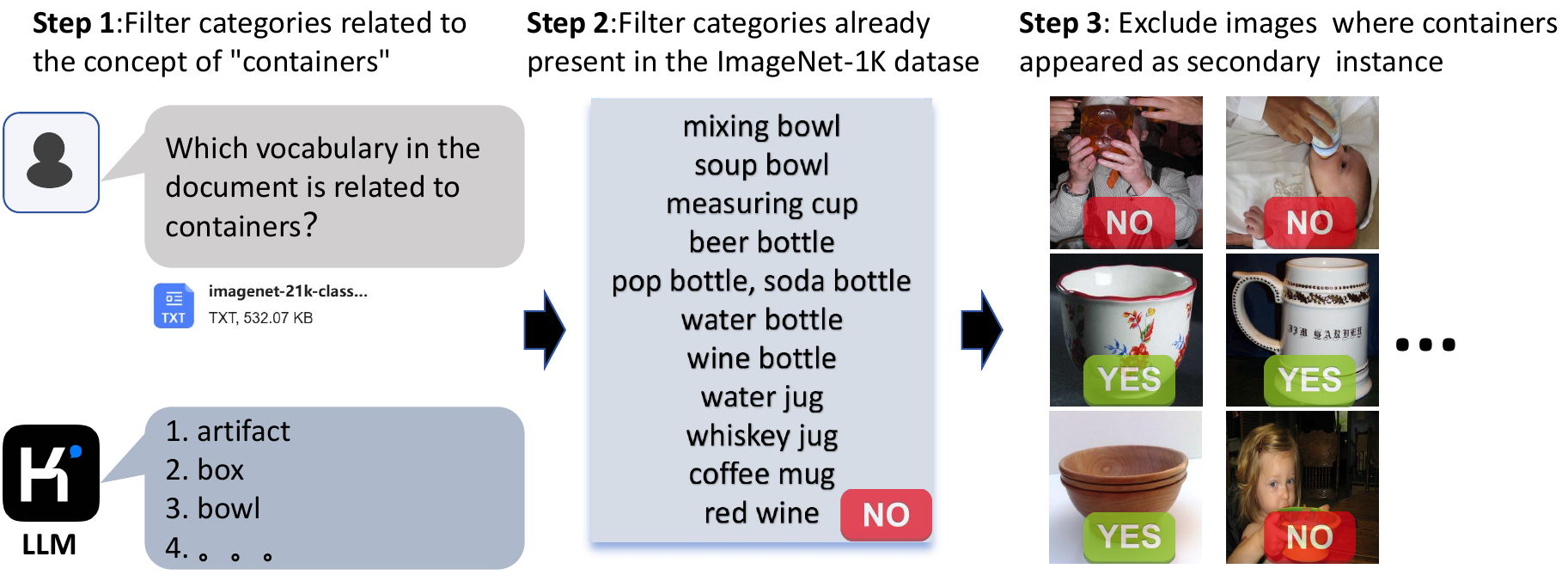}
  \caption{The detailed process for collecting container data.}
  \label{fig: container_preprocess}
\end{figure}

\subsection{Container Data}

\paragraph{Data Collection.}
As shown in Figure~\ref{fig: container_preprocess}, the process of collecting container data can be divided into three steps: \textbf{Step 1}: We employed an open-source large language model~\cite{moonshot_kimi} to sift through the 21,841 categories of ImageNet-21K, selecting those that are related to the concept of containers, such as "bowl", "cup" and others. A total of 94 categories related to the concept of containers were identified, encompassing 76399 images; \textbf{Step 2}: We removed those categories that appeared in the ImageNet-1K dataset, leaving 83 categories; \textbf{Step 3}: We removed images in which the containers are secondary instances, such as a person holding a wine glass or a baby with a bottle, leaving 51,023 container images. The statistics of the container categories is illustrated in Figure~\ref{fig: mergedDatasetInformation} (b).

\subsection{Transferred OOD Data}
\begin{figure}[t]
\centering
  \includegraphics[width=1\linewidth]{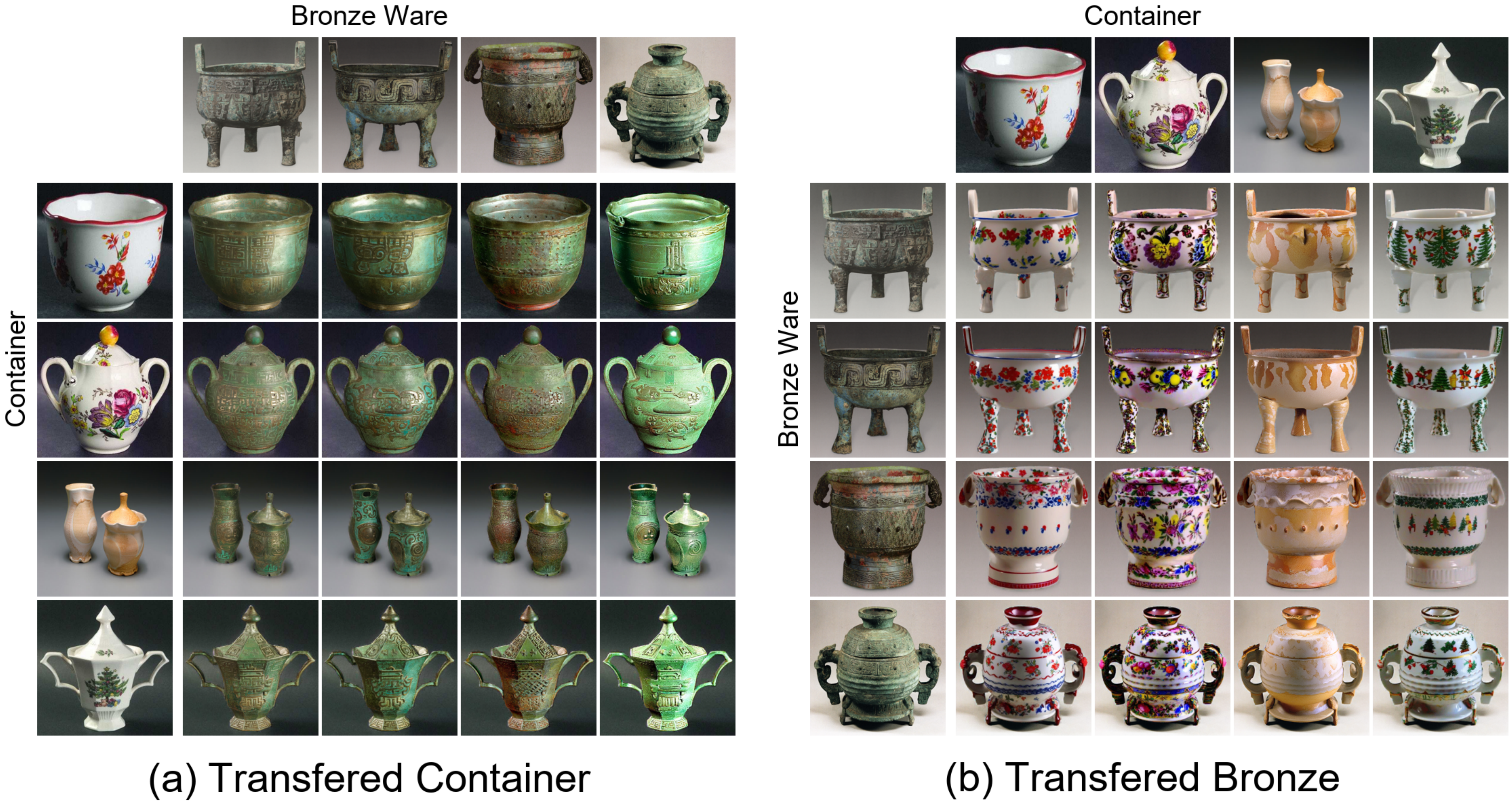}
  \caption{The generation process of transferred container and transferred bronze data using the zero-shot material transfer model ZeST~\cite{cheng2025zest}.}
  \label{fig: mergedTransfered}
\end{figure}

\paragraph{Transferred Container Data.}
By transferring the materials of bronze wares onto containers, we can obtain 51,023 transferred container data that combine the material characteristics of bronze wares with the shapes of modern containers, as shown in Figure~\ref{fig: mergedTransfered} (a). To prevent information leakage from the training data, we used 2861 bronze ware images from the test set to perform material transfers on 51023 container images, with an average of 17 container images corresponding to each bronze ware image. Consequently, the transferred container data can also be correlated with each era based on the corresponding bronze ware images. The era statistics for the transferred container data are presented in Table~\ref{tab: data Statistics}.

\vspace{-8pt}
\paragraph{Transferred Bronze Data.}
By transferring the materials of modern containers onto bronze wares, we can obtain 51,023 transferred bronze data that combine the material characteristics of containers with the shapes of bronze wares, as illustrated in Figure~\ref{fig: mergedTransfered} (b). We used 51023 container images to perform material transfers on the 2861 bronze ware test data. The statistics of transferred bronze data are consistent with those of the container data shown in Figure~\ref{fig: mergedDatasetInformation} (b).

\subsection{Distribution Shift Analysis}

To analyze the distribution of different data in ShiftedBronzes, we randomly selected 200 images from each data type: bronze ware images, transferred bronze images, container images, transferred container images, sketch and rubbing images. We used the ResNet-50~\cite{he2016deep} and ViT-B/16~\cite{dosovitskiy2021an} models pre-trained on ImageNet-1K to extract image features, then applied t-SNE for feature dimensionality reduction, and finally visualized the reduced features, as shown in Figure~\ref{fig: mergedDatasetInformation} (c). The feature gaps of sketch and rubbing images, bronze ware images, transferred bronze images, transferred container images, and container images are relatively small, indicating that their distances in the original space are close, suggesting potential difficulties in distinguishing them.

\begin{CJK*}{UTF8}{gbsn}

\section{Experiments}
\label{Experiments}

\begin{table*}[t]
\centering
\small
\caption{The comparison of seven FGVC methods on the bronze ware dataset. In addition to the overall accuracy on the test set, we have also tested each accuracy of different independent era and bronze ware category. Bold indicates the best results.}

\label{tab1:ID ACC comparsion}
\resizebox{1\textwidth}{!}{  
\begin{tabular}{cc|c|ccccccccccc}
\hline\hline
\multicolumn{2}{c|}{\multirow{3}{*}{Method}}                                                                      & \multirow{3}{*}{OA} & \multicolumn{11}{c}{Bronze Ware Category (Ding/Gui)}                                                                                                                                                                                                    \\ \cline{4-14} 
\multicolumn{2}{c|}{}                                                                                             &                     & \multicolumn{2}{c|}{Shang}                           & \multicolumn{3}{c|}{Western Zhou}                                     & \multicolumn{3}{c|}{Spring and Autumn}                                & \multicolumn{3}{c}{Warring States}               \\
\multicolumn{2}{c|}{}                                                                                             &                     & Early          & \multicolumn{1}{c|}{Late}           & Early          & Mid            & \multicolumn{1}{c|}{Late}           & Early          & Mid            & \multicolumn{1}{c|}{Late}           & Early          & Mid            & Late           \\ \hline\hline
\multicolumn{1}{c|}{\multirow{3}{*}{Single-Granularity}} & NTS-Net~\cite{NTS}                  & 73.19               & 82.35          & \multicolumn{1}{c|}{73.58}          & 79.27          & 74.58          & \multicolumn{1}{c|}{\textbf{79.55}} & 61.47          & 55.34          & \multicolumn{1}{c|}{71.31}          & 30.23          & 35.9           & 71.83          \\
\multicolumn{1}{c|}{}                                    & SPS~\cite{SPS}                    & 75.74               & 82.35          & \multicolumn{1}{c|}{79.72}          & 79.27          & \textbf{76.04} & \multicolumn{1}{c|}{76.3}           & 70.18          & 58.25          & \multicolumn{1}{c|}{68.85}          & 51.16          & 35.9           & 83.1           \\
\multicolumn{1}{c|}{}                                    & P2PNet~\cite{P2PNet}                 & 76.03               & \textbf{88.24} & \multicolumn{1}{c|}{70.05}          & \textbf{84.75} & 75.42          & \multicolumn{1}{c|}{75}             & \textbf{81.65} & 53.4           & \multicolumn{1}{c|}{\textbf{75.41}} & 48.84          & \textbf{41.03} & \textbf{88.73} \\ \hline
\multicolumn{1}{c|}{\multirow{9}{*}{Multi-Granularity}}  & \multirow{3}{*}{YourFL~\cite{YourFL}} & 99.54               & \multicolumn{11}{c}{99.60/99.45}                                                                                                                                                                                                                        \\ \cline{4-14} 
\multicolumn{1}{c|}{}                                    &                                                        & 82.37               & \multicolumn{2}{c|}{72.65}                           & \multicolumn{3}{c|}{88.17}                                            & \multicolumn{3}{c|}{79.01}                                            & \multicolumn{3}{c}{77.78}                        \\
\multicolumn{1}{c|}{}                                    &                                                        & 70.28               & 68.63          & \multicolumn{1}{c|}{73.12}          & 76.4           & 70.62          & \multicolumn{1}{c|}{68.51}          & 68.35          & 52.43          & \multicolumn{1}{c|}{62.3}           & 27.91          & \textbf{41.03} & 71.83          \\ \cline{2-14} 
\multicolumn{1}{c|}{}                                    & \multirow{3}{*}{HRN~\cite{HRN}}     & 39.57               & \multicolumn{11}{c}{25.24/62.75}                                                                                                                                                                                                                        \\ \cline{4-14} 
\multicolumn{1}{c|}{}                                    &                                                        & 85.98               & \multicolumn{2}{c|}{78.21}                           & \multicolumn{3}{c|}{90.68}                                            & \multicolumn{3}{c|}{82.62}                                            & \multicolumn{3}{c}{\textbf{83.66}}               \\
\multicolumn{1}{c|}{}                                    &                                                        & 73.85               & 78.43          & \multicolumn{1}{c|}{75.42}          & 80.18          & 75.21          & \multicolumn{1}{c|}{70.13}          & 72.02          & 54.37          & \multicolumn{1}{c|}{71.31}          & 39.53          & 23.08          & 81.69          \\ \cline{2-14} 
\multicolumn{1}{c|}{}                                    & \multirow{3}{*}{AKG~\cite{zhou2023multi}}      & \textbf{99.72}      & \multicolumn{11}{c}{\textbf{99.66/99.82}}                                                                                                                                                                                                               \\ \cline{4-14} 
\multicolumn{1}{c|}{}                                    &                                                        & \textbf{87.84}      & \multicolumn{2}{c|}{\textbf{82.76}}                  & \multicolumn{3}{c|}{\textbf{91.51}}                                   & \multicolumn{3}{c|}{\textbf{85.78}}                                   & \multicolumn{3}{c}{80.12}                        \\
\multicolumn{1}{c|}{}                                    &                                                        & \textbf{77.88}      & \textbf{88.24} & \multicolumn{1}{c|}{\textbf{81.11}} & 82.4           & 75.42          & \multicolumn{1}{c|}{\textbf{79.55}} & 74.77          & \textbf{63.11} & \multicolumn{1}{c|}{70.49}          & \textbf{53.49} & 33.33          & 84.51          \\ \hline
\end{tabular}
}
\end{table*}


\begin{table*}[t]
\centering
\caption{The comparison of eighteen OOD detection methods on the OOD data in ShiftedBronzes and five general OOD datasets. We categorized the comparative methods into three types based on their mechanisms and reported their FPR\text{@}95 and AUPOC. Bold red indicates the best performance, while underlined red denotes the second-best performance.}

\label{tab2:OOD comparsion}
\resizebox{1\textwidth}{!}{  
\begin{tabular}{c|c|c|c|c|c|c|c|c|c|c|c}
\hline\hline
Method                                               & Method Type                             &  \begin{tabular}[c]{@{}c@{}}Sketch and Rubbing\\ FPR\text{@}95${\downarrow}$ AUROC$\uparrow$ \end{tabular}           & \begin{tabular}[c]{@{}c@{}}Transfered Bronze\\ FPR\text{@}95${\downarrow}$ AUROC$\uparrow$ \end{tabular}  & \begin{tabular}[c]{@{}c@{}}Transfered Container\\ FPR\text{@}95${\downarrow}$ AUROC$\uparrow$ \end{tabular} & \begin{tabular}[c]{@{}c@{}}Container\\ FPR\text{@}95${\downarrow}$ AUROC$\uparrow$ \end{tabular}  & \begin{tabular}[c]{@{}c@{}}Species\\ FPR\text{@}95${\downarrow}$ AUROC$\uparrow$ \end{tabular}             & \begin{tabular}[c]{@{}c@{}}ImageNet-O\\ FPR\text{@}95${\downarrow}$ AUROC$\uparrow$ \end{tabular}          & \begin{tabular}[c]{@{}c@{}}iNaturalist\\ FPR\text{@}95${\downarrow}$ AUROC$\uparrow$ \end{tabular}        & \begin{tabular}[c]{@{}c@{}}Texture\\ FPR\text{@}95${\downarrow}$ AUROC$\uparrow$ \end{tabular}            & \begin{tabular}[c]{@{}c@{}}OpenImage-O\\ FPR\text{@}95${\downarrow}$ AUROC$\uparrow$ \end{tabular}         & \begin{tabular}[c]{@{}c@{}}Average\\ FPR\text{@}95${\downarrow}$ AUROC$\uparrow$ \end{tabular}     \\ \hline\hline
DICE~\cite{sun2022dice}              & \multirow{13}{*}{Post-hoc Methods}              & 98.39/26.74         & 75.11/67.79        & 94.15/38.35          & 88.68/52.48         & 84.24/57.12         & 90.93/44.59         & 93.41/31.69        & 92.60/45.89        & 93.15/41.30         & 90.07/45.11 \\ \cline{1-1} \cline{3-12} 
EBO~\cite{liu2020energy}             &                                         & 87.27/58.15         & 70.51/75.45        & 61.41/80.54          & 64.86/78.95          & 45.40/88.15         & 59.26/82.93         & 56.91/74.61        & 86.27/69.51        & 60.61/79.88         & 65.83/76.46  \\ \cline{1-1} \cline{3-12} 
GradNorm~\cite{huang2021importance}  &                                         & 89.16/57.53         & 64.02/77.46        & 65.27/79.56          & 53.92/84.05          & 38.78/90.71         & 52.93/84.63         & 52.86/77.49        & 85.72/67.14        & 57.23/81.31         & 62.21/77.76  \\ \cline{1-1} \cline{3-12} 
Gram~\cite{pmlr-v119-sastry20a}      &                                         & 83.83/28.37         & 95.24/50.25        & 87.20/61.75          & 78.52/68.92          & 80.13/77.78         & 87.14/68.33         & 74.95/65.27        & 83.76/69.40        & 85.69/64.68         & 84.05/61.64  \\ \cline{1-1} \cline{3-12} 
KL-Matching~\cite{basart2022scaling} &                                         & 95.88/43.43         & 95.47/60.14        & 96.17/52.77          & 97.43/50.18          & 96.33/65.96         & 96.30/60.98         & 97.75/41.62        & 96.66/54.15        & 96.72/55.56         & 96.52/53.87 \\ \cline{1-1} \cline{3-12} 
KNN~\cite{sun2022out}                &                                         & 89.26/58.54         & 60.06/81.07        & 45.40/86.94          & 40.93/89.39          & 17.81/95.98         & 22.09/94.61         & 18.65/94.10        & 28.04/93.03        & 22.35/94.56         & 38.29/87.58 \\ \cline{1-1} \cline{3-12} 
MDS~\cite{lee2018simple}             &                                         & 93.73/57.11         & \textcolor{red}{\underline{36.72}}/87.03        & \textcolor{red}{\textbf{22.44}}/92.20 & \textcolor{red}{\underline{14.95}}/95.17          & 10.03/96.95         & 7.11 /98.11         & 1.64 /99.48        & 1.83 /\textcolor{red}{\underline{99.61}}        & 6.88 /98.34         & 21.7/91.56 \\ \cline{1-1} \cline{3-12} 
MLS~\cite{basart2022scaling}         &                                         & 87.27/58.24         & 70.51/75.29        & 61.48/80.14          & 65.02/78.66          & 45.66/87.82         & 59.42/82.64         & 56.91/74.43        & 86.27/69.61        & 60.68/79.67         & 65.91/76.28 \\ \cline{1-1} \cline{3-12} 
MSP~\cite{hendrycks2017a}            &                                         & 88.78/60.35         & 72.15/71.39        & 68.55/71.57          & 68.55/72.17          & 54.57/80.32         & 63.76/76.17         & 64.02/69.84        & 75.56/71.23        & 63.73/74.39         & 68.85/71.94 \\ \cline{1-1} \cline{3-12} 
ODIN~\cite{liang2018enhancing}       &                                         & 55.34/85.02         & 80.87/72.90        & 77.52/81.14          & 88.52/82.01          & 80.48/82.88         & 94.50/74.69         & 79.61/81.23        & 96.40/71.88        & 87.91/79.27         & 82.35/79.0 \\ \cline{1-1} \cline{3-12} 
OpenMax~\cite{bendale2016towards}    &                                         & 89.84/62.33         & 73.41/69.93        & 70.00/72.64          & 70.03/69.85          & 55.50/79.80         & 64.86/76.79         & 65.24/69.78        & 74.60/71.81        & 64.69/74.91         & 69.8/71.98 \\ \cline{1-1} \cline{3-12} 
ReAct~\cite{sun2021react}            &                                         & 87.30/54.94         & 64.89/77.34        & 63.76/78.80          & 65.18/78.66          & 47.52/87.17         & 60.55/81.34         & 58.97/72.51        & 87.78/68.41        & 61.32/78.42         & 66.36/75.29 \\ \cline{1-1} \cline{3-12} 
VIM~\cite{wang2022vim}               &                                         & 86.43/55.05         & 38.78/\textcolor{red}{\underline{90.60}}        & 22.77/\textcolor{red}{\textbf{94.89}} & 17.94/\textcolor{red}{\underline{96.30}}          & \textcolor{red}{\underline{4.47}}/\textcolor{red}{\underline{98.97}}         & \textcolor{red}{\underline{4.47}}/\textcolor{red}{\underline{98.96}}         & \textcolor{red}{\underline{1.41}}/\textcolor{red}{\underline{99.72}}        & \textcolor{red}{\underline{1.29}}/99.59        & \textcolor{red}{\underline{4.50}}/\textcolor{red}{\underline{99.03}}         & \textcolor{red}{\underline{20.23}}/\textcolor{red}{\underline{92.57}} \\ \hline\hline
OpenGAN~\cite{kong2021opengan}       & \multirow{2}{*}{Generation-based Methods} & 50.44/81.32         & 79.18/58.59        & 45.5/78.24           & 43.36/78.25          & 45.74/72.27         & 48.55/74.51         & 39.33/81.56        & 71.89/73.95        & 45.78/76.9          & 52.2/75.07 \\ \cline{1-1} \cline{3-12} 
DIFFGUARD~\cite{gao2023diffguard}    &                                         & \textcolor{red}{\underline{37.4}}/\textcolor{red}{\underline{87.37}}         & 97.93/34.6         & 97.16/40.22          & 98.07/19.51           & 99.59/7.69          & 86.84/47.16         & 61.85/72.54        & 70.97/61.25        & 95.19/30.07         & 82.78/44.49 \\ \hline\hline
LoCoOp~\cite{miyai2024locoop}        & \multirow{3}{*}{VLM-based Methods}             & 53.72/84.71         & 47.94/86.24        & 64.29/74.86          & 41.85/86.77          & 61.97/82.67         & 41.83/88.19         & 90.44/53.91        & 20.0/96.16         & 61.06/76.48         & 53.68/81.11 \\ \cline{1-1} \cline{3-12} 
ID-like~\cite{bai2024id}             &                                         & \textcolor{red}{\textbf{24.55}}/\textcolor{red}{\textbf{94.94}} & \textcolor{red}{\textbf{5.13}}/\textcolor{red}{\textbf{98.8}} & \textcolor{red}{\underline{22.72}}/\textcolor{red}{\underline{94.43}}          & \textcolor{red}{\textbf{0.55}}/\textcolor{red}{\textbf{99.79}} & \textcolor{red}{\textbf{0.07}}/\textcolor{red}{\textbf{99.93}} & \textcolor{red}{\textbf{0.03}}/\textcolor{red}{\textbf{99.95}} & \textcolor{red}{\textbf{0.0}}/\textcolor{red}{\textbf{99.99}} & \textcolor{red}{\textbf{0.0}}/\textcolor{red}{\textbf{99.98}} & \textcolor{red}{\textbf{0.23}}/\textcolor{red}{\textbf{99.92}} & \textcolor{red}{\textbf{5.92}}/\textcolor{red}{\textbf{98.64}}  \\ \cline{1-1} \cline{3-12} 
CLIPN~\cite{wang2023clipn}           &                                         & 58.55/82.87          & 48.82/80.61        & 69.65/69.34          & 7.75/97.72          & 7.25/98.11          & 7.71/98.12          & 21.82/94.26        & 8.46/97.59         & 13.53/96.25         & 27.06/90.54 \\ \hline
\end{tabular}
}
\end{table*}

\begin{figure*}[tb]
\centering
  \includegraphics[width=1\linewidth]{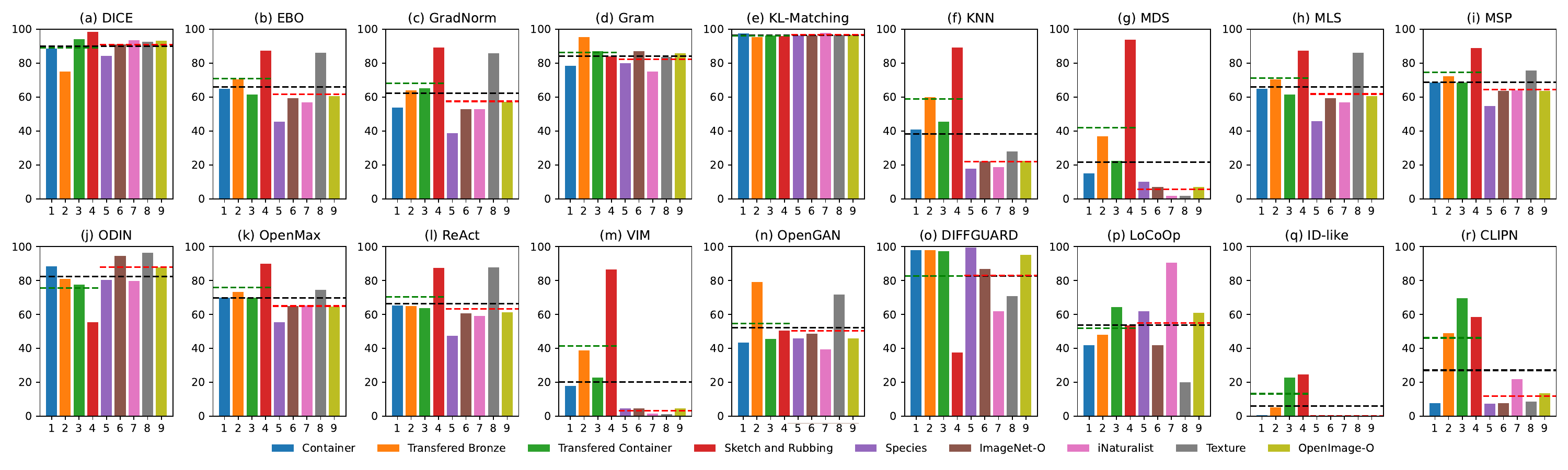}
  \caption{The comparison of FPR\text{@}95 performance of 18 OOD detection methods on the OOD data in ShiftedBronzes and five general OOD datasets. The black dashed line represents the average performance across all OOD data. The green dashed line indicates the average performance on the \textit{hard OOD data} (container, transferred bronze, transferred container, sketch and rubbing). The red dashed line indicates the average performance on the \textit{easy OOD data} (Species, ImageNet-O, iNaturalist, Texture, OpenImage-O).}
  \label{fig: rsult_bar_image}
\end{figure*}

\begin{figure*}[tb]
\centering
  \includegraphics[width=1\linewidth]{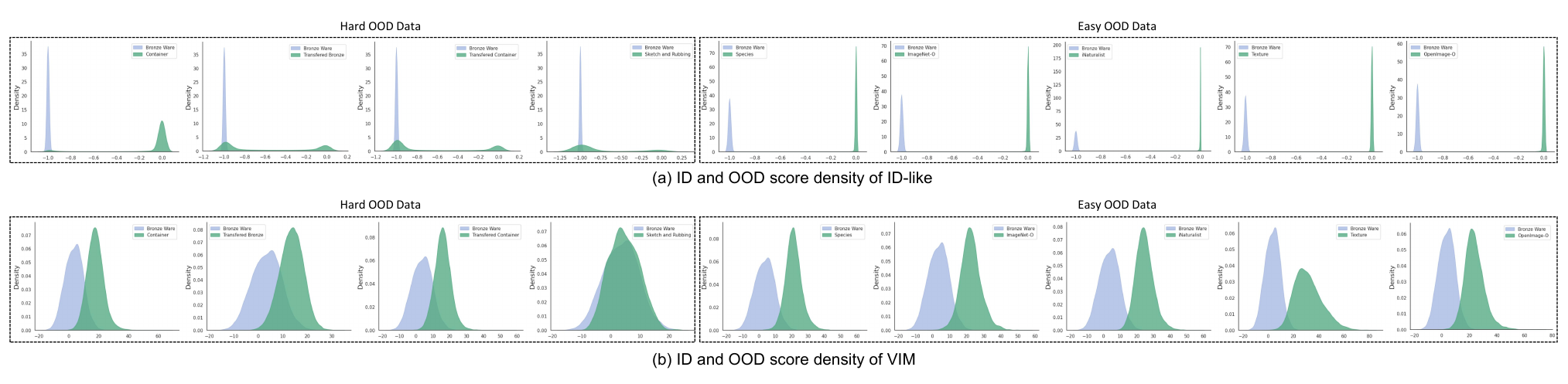}
  \caption{ID and OOD score density distribution of two top-performing methods on the \textit{hard OOD data} and \textit{easy OOD data}.}
  \label{fig: distribution_density}

\end{figure*}

\subsection{Data Preparation}
\label{Data Preparation}

The bronze ware data from our proposed ShiftedBronzes is used as ID data to train the bronze ware dating methods. Following the divisions of bronze Ding dataset~\cite{zhou2023multi} and other fine-grained classification datasets~\cite{zhou2023multi, Large-scale-Fashion}, we split the 5711 bronze Ding and Gui images into three sets: the scales of the training set, validation set, and test set are 4:1:5 (2278:572:2861), respectively. And each fine-grained period and bronze ware category is split following this ratio.

Besides the ShiftedBronzes OOD data, we additionally utilized five general OOD datasets from ImageNet-1K, including Species~\cite{basart2022scaling}, ImageNet-O~\cite{hendrycks2021natural}, iNaturalist~\cite{huang2021mos}, Texture~\cite{kylberg2011kylberg}, and OpenImage-O~\cite{wang2022vim}. Considering the degree of distribution shift from the ID data, we classified the seven OOD data types in ShiftedBronzes as \textit{hard OOD data} and the five general OOD datasets as \textit{easy OOD data}.

\subsection{Implementation Details}
\paragraph{Devices and Code.} All experiments were implemented by PyTorch, and conducted on a server with 4 RTX A40 GPUs and Intel$\circledR$ Xeon$\circledR$ Gold 5220 CPUs (72 cores). The code for all post-hoc OOD detection methods and OpenGAN method are derived from openOOD benchmark~\cite{yang2022openood}. And the code for other methods are derived from the corresponding official repositories. The hyperparameters for each method are set according to the optimal configurations reported in their respective papers.

\vspace{-8pt}
\paragraph{Evaluation.} For evaluation metrics, we employ FPR@95 and AUROC for the OOD detection task and OA and accuracy for the dating task. AUROC is a metric that computes the area under the receiver operating characteristic curve. A higher value indicates better detection performance. FPR@95 is short for FPR@TPR95, which is the false positive rate when the true positive rate is 95\%. Smaller FPR@95 implies better performance. We evaluate the overall performance of different bronze ware dating methods using overall accuracy (OA) and assess their individual performance at various levels of granularity using independent accuracy.

\subsection{Bronze Ware Dating}
\paragraph{Setting.}
We select five representative FGVC methods and one bronze ware dating method to evaluate their performance on the ShiftedBronzes dataset. Based on the experimental settings in \cite{zhou2023multi}, we chose three multi-granularity FGVC methods (YourFL~\cite{YourFL}, HRN~\cite{HRN}, AKG~\cite{zhou2023multi}) and three single-granularity FGVC methods (NTS-Net~\cite{NTS}, SPS~\cite{SPS}, P2PNet~\cite{P2PNet}) for the bronze ware dating task. Notably, AKG~\cite{zhou2023multi} is specifically designed for the dating of bronze artifacts

\vspace{-6pt}
\paragraph{Results and Analysis.}
(1) \textit{Domain-specific expert knowledge significantly enhances the model learning for bronze ware dating.} As can be seen from Table~\ref{tab1:ID ACC comparsion}, the AKG achieves the best OA performance. Moreover, it also attains the best independent accuracy across two bronze ware categories, three coarse-granularity eras and five fine-granularity eras. Designed with the dating of bronze Ding, the AKG outperform other general FGVC methods even when applied to the expanded data with Gui. (2) \textit{Multi-level feature enhancement does not always yield effective improvements.} Except for the top-performing AKG, the other two multi-granularity FGVC methods underperform compared to SPS and P2PNet. And the performance of HRN on bronze ware category classification is significantly lower than that of other multi-granularity methods. This indicates that merely obtaining additional information from hierarchical labels is insufficient, and multi-granularity features may interfere with each other.

\vspace{-5pt}
\subsection{OOD Detection Experiments}
\subsubsection{Setting}
We selected eighteen widely studied OOD detection methods, which are divided into three categories (post-hoc, VLM-based, generation-based), and conducted OOD detection experiments on ShiftedBronzes dataset, as well as five general OOD datasets. The post-hoc methods include DICE~\cite{sun2022dice}, EBO~\cite{liu2020energy}, GradNorm~\cite{huang2021importance}, Gram~\cite{pmlr-v119-sastry20a}, KL-Matching~\cite{basart2022scaling}, KNN~\cite{sun2022out}, MDS~\cite{lee2018simple}, MLS~\cite{basart2022scaling}, MSP~\cite{hendrycks2017a}, ODIN~\cite{liang2018enhancing}, OpenMax~\cite{bendale2016towards}, ReAct~\cite{sun2021react} and VIM~\cite{wang2022vim}. We selected the AKG, which demonstrated the best performance in the bronze ware dating experiments, to serve as the pre-model for post-hoc methods.

The VLM-based methods include ID-like~\cite{bai2024id}, LoCoOp~\cite{miyai2024locoop} and CLIPN~\cite{wang2023clipn}. ID-like ~\cite{bai2024id} and LoCoOp~\cite{miyai2024locoop} employ few-shot prompt learning for training. We conducted experiments to evaluate their performance under varying numbers of training samples and selected the best-performing configurations to compared with other methods. CLIPN~\cite{wang2023clipn} has been pre-training on the CC3M~\cite{sharma2018conceptual} dataset, followed by zero-shot inference.

The generation-based methods include OpenGAN~\cite{kong2021opengan} and DIFFGUARD~\cite{gao2023diffguard}. The pre-model required by OpenGAN uses the same AKG as the post-hoc methods.

\begin{figure*}[tb]
\centering
  \includegraphics[width=1\linewidth]{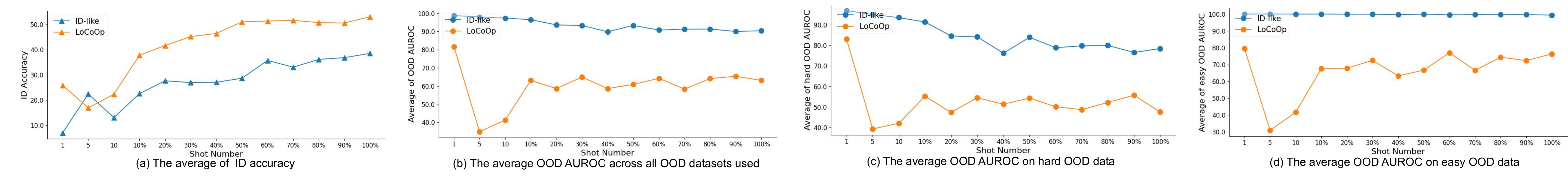}
  \caption{Comparison of ID and OOD performance of ID-like~\cite{bai2024id} and LoCoOp~\cite{miyai2024locoop} under varying amounts of training data.}
  \label{fig: few_shot_ID_result}
\end{figure*}

\begin{table}[hb]
\centering
\caption{Comparison of the average performance of post-hoc, VLM-based and generation-based methods.}
\label{tab_data_comprasion_of_hard_and_easy}
\resizebox{0.8\linewidth}{!}{  
\begin{tabular}{c|c|c|c}
\hline\hline
                 & \begin{tabular}[c]{@{}c@{}}All OOD Data\\ FPR\text{@}95${\downarrow}$ AUROC$\uparrow$\end{tabular} & \begin{tabular}[c]{@{}c@{}}Hard OOD Data\\ FPR\text{@}95${\downarrow}$ AUROC$\uparrow$\end{tabular} & \begin{tabular}[c]{@{}c@{}}Easy OOD Data\\ FPR\text{@}95${\downarrow}$ AUROC$\uparrow$\end{tabular} \\ \hline\hline
Post-hoc         & 64.01/73.93                                                & 70.79/69.82                                                 & 58.60/77.21                                                 \\ \hline
VLM-based       & \textbf{28.89/90.10}                                       & \textbf{37.13/87.59}                                        & \textbf{22.29/92.10}                                        \\ \hline
Generation-based & 67.49/59.78                                                & 68.63/59.76                                                 & 66.57/59.79                                                 \\ \hline
\end{tabular}
}
\end{table}

\subsubsection{Results and Analysis}

As shown in the red-marked metrics in Table 1, ID-like~\cite{bai2024id} and DIFFGUARD~\cite{gao2023diffguard} achieved the top-2 average FPR@95 and AUROC performance on the sketch and rubbing data. On the transferred and container data, ID-like, MDS~\cite{lee2018simple}, and VIM~\cite{wang2022vim} dominated the top-2 average FPR@95 and AUROC performance. For the general data, ID-like and VIM largely dominated the top-2 performances. Through further analysis of these methods across various OOD datasets, we derived some empirical findings, which we hope will contribute to the advancement of OOD detection algorithms for domain-specific data.

(1) \textit{Pre-trained diffusion model-based methods are sensitive to color distribution shift.} DiffGuard and OpenGAN did not perform well on most of the other datasets. However, DiffGuard achieved performance just below ID-like~\cite{bai2024id} on the sketch and rubbing data, highlighting the potential of pre-trained diffusion models for detecting OOD samples with color distribution shifts. Analyzing the sensitivity of generation-based methods to color distribution shifts may provide insights for improving their performance on other datasets.

(2) \textit{General knowledge in VLMs and proprietary knowledge in specialized datasets are key to VLM-based methods outperforming others.} VLMs are pre-trained in an open-world setting, which provides VLM-based methods with an inherent advantage in OOD detection. As shown in Figure~\ref{tab_data_comprasion_of_hard_and_easy}, VLM-based methods outperform all others in terms of average performance across various OOD datasets and difficulty levels. Among the three VLM-based methods tested, ID-like~\cite{bai2024id} consistently achieved the best or second-best performance across all OOD data and average metrics. In contrast, CLIPN~\cite{wang2023clipn} and LoCoOp~\cite{miyai2024locoop}, although also based on VLMs, did not show any significant advantage. This can be attributed to differences in their prompt architectures: ID-like incorporates both learnable ID and OOD prompts, allowing it to leverage expert knowledge from ID samples. On the other hand, LoCoOp and CLIPN only have learnable OOD prompts, which limits their ability to capture ID-specific information. Furthermore, CLIPN’s pretraining on the CC3M dataset~\cite{sharma2018conceptual}, which lacks domain-specific bronze ware data, further hinders its performance in bronze ware OOD detection.

(3) \textit{ID and OOD performance variations during fine-tuning VLM-based methods with varying sample sizes differ across specialized and general domains.} We tested the ID and OOD performance of ID-like~\cite{bai2024id} and LoCoOp~\cite{miyai2024locoop} methods with varying training sample sizes, as shown in Figure~\ref{fig: few_shot_ID_result}. Ming et al.\cite{ming2024does} examined the effect of training sample size on the Caltech-101 dataset, finding that both ID and OOD performance improved as the sample size increased. However, our results on the ShiftedBronzes benchmark reveal a different trend. Both ID-like and LoCoOp demonstrated their best OOD detection performance with the 1-shot training paradigm. As the number of samples for fine-tuning increased, performance on \textit{easy OOD data} with larger distribution shifts stabilized, while performance on \textit{hard OOD data} with smaller shifts declined. This indicates that OOD detection in specialized domains, such as our proposed ShiftedBronzes, exhibits distinct patterns compared to general domain OOD detection, possibly due to more domain-specific characteristics.

(4) \textit{Trainable ID prompts make the performance of VLM-based methods more stable.} Although the overall trend of LoCoOp~\cite{miyai2024locoop} is similar to that of ID-like~\cite{bai2024id} when trained with different numbers of samples, it exhibits significant fluctuations, particularly when fewer samples are used. As shown in Figure~\ref{fig: few_shot_ID_result}(b-d), the performance of LoCoOp fluctuates noticeably as the number of training samples increases, with more pronounced fluctuations when fewer samples are used. In contrast, ID-like, which trains both ID and OOD prompts, shows more stable performance than LoCoOp, which trains only the OOD prompt. This offers insights for the design of VLM-based OOD detection methods.

(5) \textit{Post-hoc methods that leverage diverse features throughout the pipeline, as well as those that can adaptively tune parameters of score function based on specialized domain data, often demonstrate superior performance.} Although post-hoc methods generally underperform VLM-based methods, they offer greater flexibility and ease of deployment. Analyzing these methods is crucial for developing future approaches that may surpass VLM-based methods. As shown in Table~\ref{tab2:OOD comparsion}, among post-hoc methods, MDS~\cite{lee2018simple} and ViM~\cite{wang2022vim} exhibit superior performance. MDS computes OOD scores using Mahalanobis distances and allows parameters of score function adjustment based on ID samples, which enhances robustness across different OOD datasets. ViM, on the other hand, combines softmax scores, logits, and residuals from the feature space to compute OOD scores, outperforming methods that rely on a single source of input, such as features (KNN~\cite{sun2022out}), max logits (MLS~\cite{basart2022scaling}), or max softmax (MSP~\cite{hendrycks2017a}).

(6) \textit{Compared to general OOD data, the smaller distribution shift in domain-specific OOD data poses a greater challenge for current OOD detection methods.} 
Compared to \textit{easy OOD data}, \textit{hard OOD data} exhibits a smaller distribution shift relative to ID data, a pattern also observed in other specialized domains. To explore the impact of distribution shifts on OOD detection performance, we compare the performance of various methods on both hard and \textit{easy OOD data}. As shown in Figure~\ref{fig: rsult_bar_image}, 13 out of the 18 methods evaluated show weaker average FRP@95 on \textit{hard OOD data} than on \textit{easy OOD data}. The top five methods in terms of average FRP@95 perform significantly worse on \textit{hard OOD data} compared to \textit{easy OOD data}. Notably, ID-like~\cite{bai2024id} and ViM~\cite{wang2022vim} demonstrate the top-2 performance among the 18 methods. To further investigate, we plot the ID and OOD score density distributions for these two top-performing methods on the bronze ware test set, as well as on hard and \textit{easy OOD data}, shown in Figure~\ref{fig: distribution_density}. The score density distributions for ID and \textit{hard OOD data} exhibit significant overlap, while the distinction is clearer on the \textit{easy OOD data}. These results suggest that top-performing OOD detection methods struggle with OOD samples that exhibit small distribution shifts relative to the ID data.

\end{CJK*}

\section{Limitations and Conclusion}
There are several limitations in our study that should be addressed in future work. First, the lack of data from other specialized domains may limit the generalizability of our findings. Additionally, our comparison does not cover all types of OOD detection methods, which could affect the comprehensiveness of our conclusions.

In this study, we introduce a novel benchmark ShiftedBronzes to assess the dating of bronze wares amidst diverse domain distribution shifts within open-world scenarios. With a thorough comparison of six FGVC and eighteen OOD detection methods, we demonstrate that bronze ware dating with domain distribution shifts remains a challenge and requires further attention from the research community. Furthermore, we empirically investigate how OOD performance in specialized domain is influenced by various factors, including distribution shift types, detector architecture, pre-training, etc. We hope our research can contribute to both the archaeology and the development of OOD detection algorithms.

{
    \small
    \bibliographystyle{ieeenat_fullname}
    \bibliography{main}
}

\end{document}